\documentclass{article}
\usepackage{ijcai21}

\usepackage{times}
\usepackage{soul}
\usepackage{url}
\usepackage[hidelinks]{hyperref}
\usepackage[utf8]{inputenc}
\usepackage[small]{caption}
\usepackage{graphicx}
\usepackage{amsmath}
\usepackage{amsthm}
\usepackage{booktabs}
\usepackage{algorithm}
\usepackage{algorithmic}
\usepackage[T1]{fontenc}
\usepackage{color}
\usepackage{amsfonts}
\usepackage{multirow}
\usepackage{booktabs}

\usepackage{marvosym}
\usepackage{amssymb}
\usepackage{multirow}
\usepackage{threeparttable}
\usepackage{booktabs}
\usepackage{hyperref}
\usepackage{subfigure}
\usepackage{algorithm,algorithmic,paralist}
\usepackage{bm}

\title{Learning from Crowds with Sparse and Imbalanced Annotations}

\author{
	Ye Shi$^1$\and
	Shao-Yuan Li\footnote{This research was supported by National Natural Science Foundation of China (61906089), Jiangsu Province Basic Research Program (BK20190408), and China Postdoc Science Foundation (2019TQ0152). Shao-Yuan Li is the corresponding author.}$^{1,2}$ \and  
	Sheng-Jun Huang$^{1}$ 
	\affiliations
	$^1$Ministry of Industry and Information Technology Key Laboratory of Pattern Analysis and Machine Intelligence College of Computer Science and Technology, Nanjing University of Aeronautics and Astronautics, Nanjing, 211106, China\\
	$^2$State Key Laboratory for Novel Software Technology, Nanjing University, Nanjing, 210023, China
	\emails
	\{shiye1998, lisy, huangsj\}@nuaa.edu.cn
}

\begin{document}

\maketitle

\begin{abstract}

Traditional supervised learning requires ground truth labels for the training data, whose collection can be difficult in many cases. Recently, crowdsourcing has established itself as an efficient labeling solution through resorting to non-expert crowds. To reduce the labeling error effects, one common practice is to distribute each instance to multiple workers, whereas each worker only annotates a subset of data, resulting in the {\it sparse annotation} phenomenon. In this paper, we note that when meeting with class-imbalance, i.e., when the ground truth labels are {\it class-imbalanced}, the sparse annotations are prone to be skewly distributed, which thus can severely bias the learning algorithm. To combat this issue, we propose one self-training based approach named {\it Self-Crowd} 
by progressively adding confident pseudo-annotations and rebalancing the annotation distribution. Specifically, 
we propose one distribution aware confidence measure to select confident pseudo-annotations, which adopts the resampling strategy to oversample the minority annotations and undersample the majority annotations. On one real-world crowdsourcing image classification task, we show that the proposed method yields more balanced annotations throughout training than the distribution agnostic methods and substantially improves the learning performance at different annotation sparsity levels.
\end{abstract}

\section{Introduction}

The achievements of deep learning rely on large amounts of ground truth labels, but collecting them is difficult in many cases. To alleviate this problem, crowdsourcing provides a time- and cost-efficient solution through collecting non-expertise annotations from crowd workers. Learning from training data with crowdsourcing annotations is called learning from crowds or crowdsourcing learning, and has attracted much attention during the last years\cite{buecheler2010crowdsourcing}. 

As the crowds can make mistakes, one core task in crowdsourcing learning is to deal with the annotation noise, for which purpose many approaches have been proposed\cite{DS,Raykar10,DennyZhouNips12,crowdlayer}. In this paper, we move one step further by noticing the sparsity and class-imbalance phenomenon of crowdsourcing annotations in real-world applications. 
\begin{figure}[tp]
	\centering
	\includegraphics[width=\columnwidth]{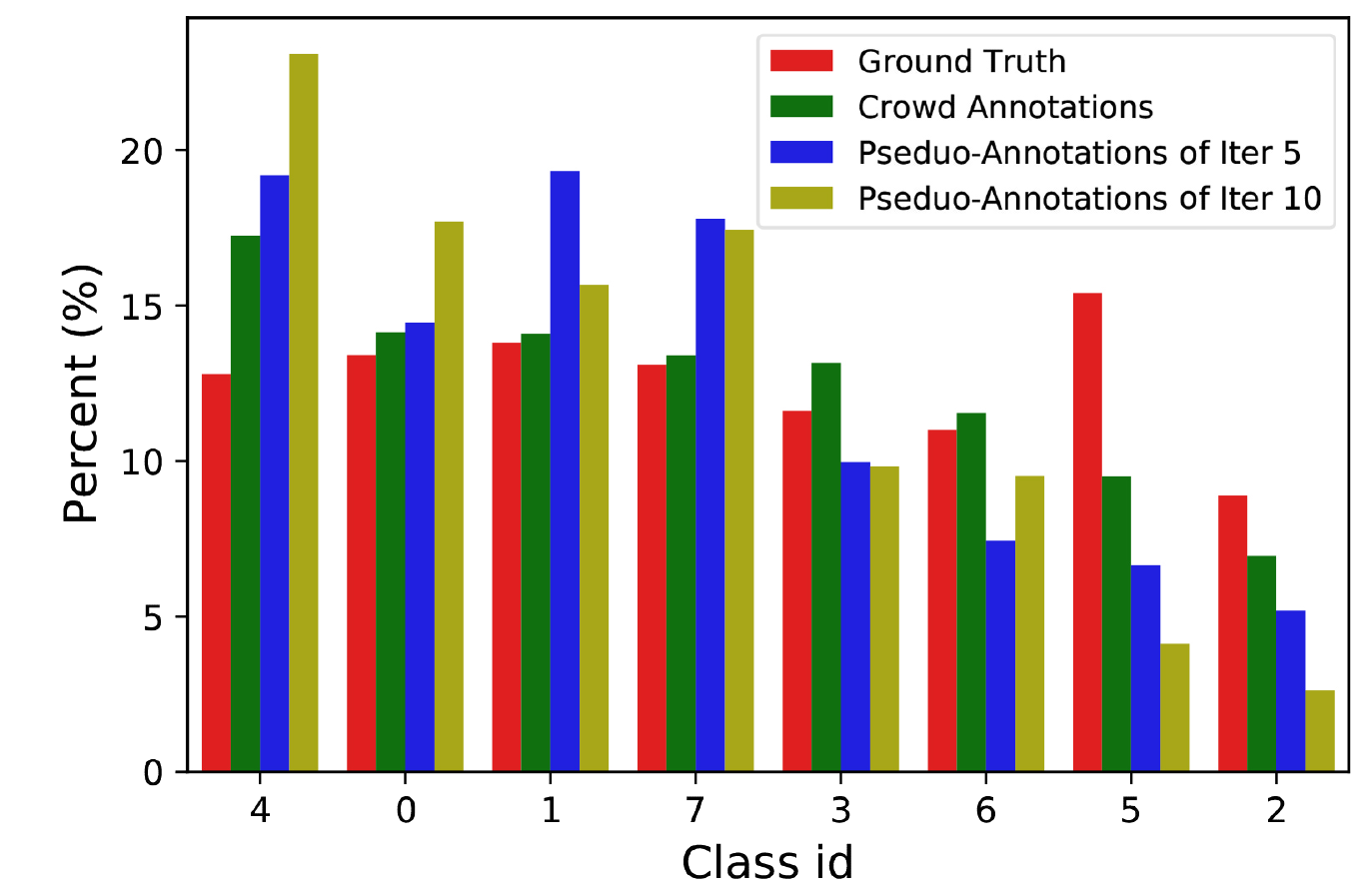}
	\caption{Class distributions of ground truth labels, observed annotations and two intermediate steps (iteration 5 and 10) of confidence based self-training. In each iteration, $10,000$ pseudo-annotations are added into the training data.}
	\label{fig:hist_classes}
\end{figure}

In crowdsourcing, annotation sparsity is common. For example, to reduce the labeling error effects, repetitive labeling is employed to introduce labeling redundancy, i.e., each instance is distributed to more than one workers. At the same time, to collect annotations efficiently, a rather large number of workers are employed, whereas each worker only annotates a subset of data. This results in sparse annotations. We note that when meeting with class-imbalance, i.e., when the ground truth labels of the concerned task are class-imbalanced, the sparsity can lead to inevitable skewed distribution, which may severely bias the learning algorithm.

Here we show one real-world example. LabelMe\cite{labelme} is an image crowdsourcing dataset, consisting of $1000$ training data with annotations collected from $59$ workers through the Amazon Mechanical Turk (AMT) platform. On average, each image is annotated by $2.547$ workers, and each worker is assigned with $43.169$ images. This sparsity on one hand makes estimating each worker's expertise quite challenging. On the other hand, Figure \ref{fig:hist_classes} shows the effects of sparsity encountering class-imbalance. Except for the ground truth labels and observed crowd annotations, we also show results of two intermediate steps for normal confidence based self-training, i.e., the most confident pseudo-annotations are iteratively added into the training data for updating the model. For the ground truth labels and observed annotations, their standard deviations over classes are respectively $1.85\%$ and $2.95\%$, meaning a more skewed annotation distribution. Moreover, the skewness has biased the self-training algorithm to prefer majority class annotations and ignore the minority classes, which in turn leads to more severely skewed annotations and learning bias. We will show in the experiment section that this bias significantly hurts the learning performance. Nevertheless, this issue has been rarely paid attention to and touched in crowdsourcing learning. 


In this paper, we propose one distribution aware self-training based approach to combat this issue. At a high level, we iteratively add confident pseudo-annotations and rebalance the annotation distribution. Within each iteration, we efficiently train a deep learning model using available annotations, and then use it as a teacher model to generate pseudo-annotations. To alleviate the imbalance issue, we propose to select the most confident pseudo-annotations using resampling strategies, i.e., we undersample the majority classes and oversample minority classes. Then the learning model is retrained on the combination of observed and pseudo-annotations. We name our approach Self-Crowd, and empirically show its positive effect at different sparsity levels on the LabelMe dataset.

\section{The Self-Crowd Framework}
With $\mathcal{X} \subset \mathbb{R}^d$ denoting feature space and $\mathcal{Y} = \{1,2,\cdots,C\}$ denoting label space, we use $x \in \mathcal{X}$ to denote the instance, as well as $y, \overline{y} \in \mathcal{Y}$ denote the corresponding ground truth labels and crowd annotations respectively. Let $D = \{(x_i, {\overline{y}_i})\}^{N}_{i=1}$ denote the training data with $N$ instances, and $\overline{y}_i = \{{\overline{y}^{r}_{i}}\}_{r=1}^{R}$ denote the crowdsourcing annotations from $R$ workers.

In the following, we first introduce the deep crowdlayer model proposed by \cite{crowdlayer} as our base learning model, then propose the distribution aware confidence measure to deal with annotation sparsity and class-imbalance, and finally summarize the algorithm procedure.

\subsection{Deep Crowdlayer Base Learning Model}
\begin{figure}[tp]
	\centering
	\includegraphics[width=\columnwidth]{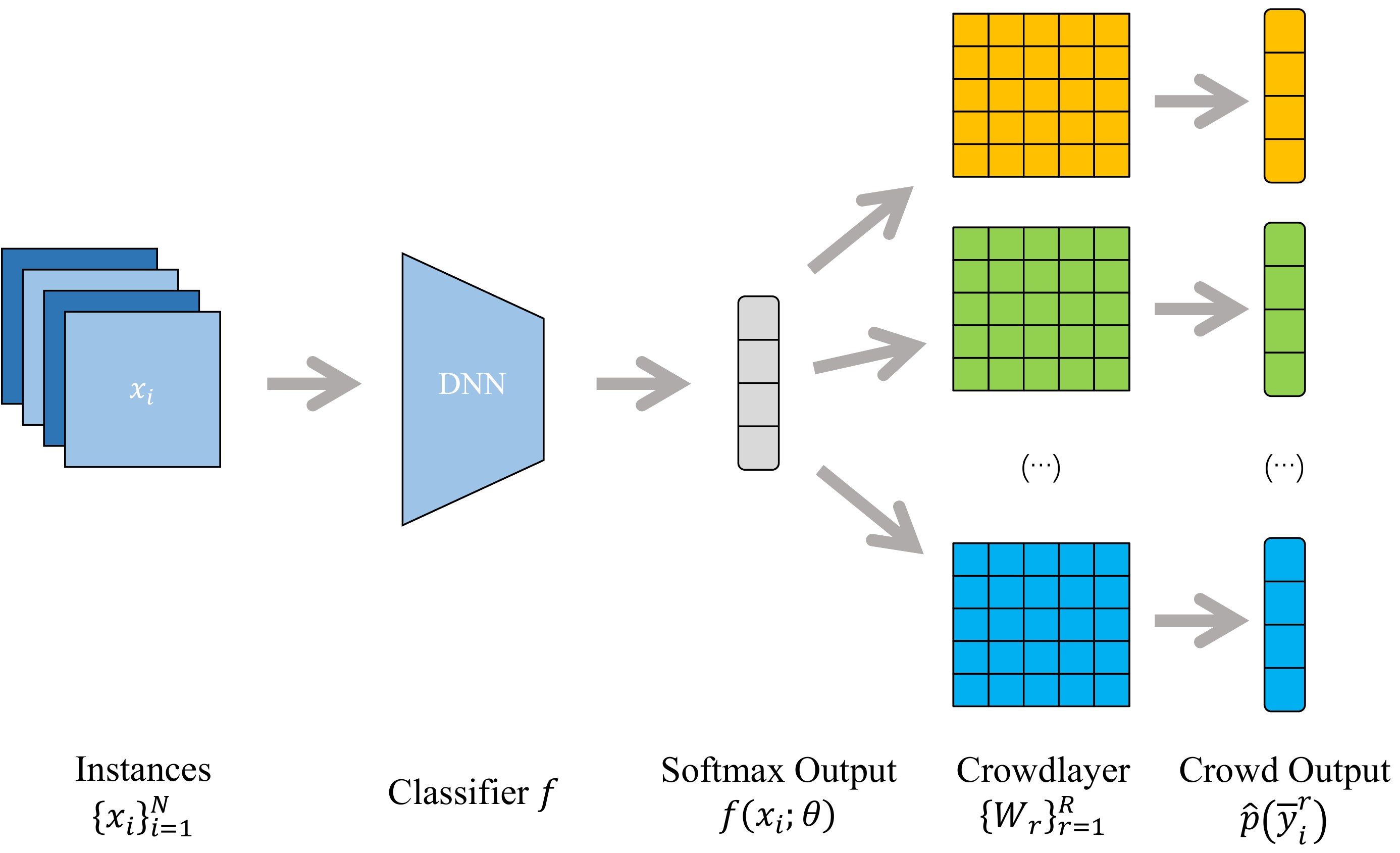}
	\caption{The network architecture of the deep crowdlayer model.}
	\label{fig:model_arc}
\end{figure}

With the ubiquitous success of deep neural networks (DNN), deep crowdsourcing learning has been studied by combining the strength of DNN with crowdsourcing. As one of the pioneers in this direction, \cite{aggnet} extended the classic DS model\cite{DS} by using a convolutional neural network (CNN) classifier as the latent true label prior, and conducted optimization using expectation-maximization (EM) algorithm. To avoid the computational overhead of the iterative EM procedure, \cite{crowdlayer} introduced the crowdlayer model and conducted efficient end-to-end SGD optimization.

In detail, using $f(x;\theta)\in [0,1]^{1 \times C}$ to denote the softmax output of the deep neural network classifier $f(\cdot)$ with parameter $\theta$ for some instance $x$, the crowdlayer model introduced $R$ parameters $\{W_r \in \mathbb{R}^{C \times C}\}_{r=1,\cdots,R}$ to capture the annotating process of the crowds, i.e., the annotations of $x$ given by worker $r$ is derived as:
\begin{equation}\label{annpred}
	\hat{p}(\overline{y}^r) = softmax(f(x;\theta) \cdot W_r).
\end{equation}

While $W_r$ is real valued without any structural constraints, it can be used to represent the worker's annotating expertise, i.e., $W_r(i,j)$ can denote the process that instances belonging to class $i$ are annotated with class label $j$ by worker $r$. Larger diagonal values mean better worker expertise. Given a specific loss function $\ell$, e.g., the cross entropy loss used in this paper, the loss over the crowdsourcing training data $D$ is defined as:
\begin{eqnarray}\label{loss} 	
	L:= \sum_{i=1}^{N}\sum_{r=1}^{R} \mathcal{I}[\overline{y}^{r}_{i}\neq 0] \ell(\hat{p}(\overline{y}^{r}_{i}), \overline{y}^{r}_{i}).
\end{eqnarray}
Here $\mathcal{I}$ is the indicator function. Then regarding $W_r$ as one crowdlayer after the neural network classifier $f(\cdot)$, \cite{crowdlayer} proposed to simultaneously optimizing the classifier parameter $\theta$ and $W_r$ in an end-to-end manner by minimizing the loss defined in Eq.~\ref{loss}.

The network architecture of the crowdlayer is shown in Figure~\ref{fig:model_arc}. Actually, this architecture and the end-to-end loss optimization over the derived loss in Eq.~\ref{loss} have been the cornerstone of various deep crowdsourcing learning approaches\cite{annoReg,commonConfusions,SpeeLFC}. They mainly differ in specific structural regularization over the expertise parameters $W_r$ with different motivations. In this paper, we adopt the straightforward crowdlayer as our base learning model for simplicity, and focus on using the self-training idea to deal with the sparse and class-imbalanced issue in crowdsourcing annotations.
  
\subsection{Distribution Aware Confidence Measure}
To combat the annotation sparsity and class-imbalance, we use the crowdlayer model as the base model, and propose one distribution aware confidence measure to conduct self-training. During the training, we progressively predict pseudo-annotations for the unannotated instances for each worker, and add some of them into the training data, then update the learning model. The most confident pseudo-annotations which contribute to rebalancing the annotation distribution are selected. Next, we will explain the measure in detail.

\begin{algorithm}[tp]
	\caption{The Self-Crowd Framework}
	\label{alg:algorithm}
	\begin{algorithmic}[1]
		\STATE \textbf{Input}: 
		\STATE $D = \{(x_i, {\overline{y}_i})\}^{N}_{i=1}$ : crowdsourcing training data
		\STATE $\ell$: loss function
		\STATE \textbf{Output}: classifier $f$
		\STATE \textbf{Initialization:}
		\STATE train the crowdlayer model using the loss in Eq.~\ref{loss} on $D$

		\STATE obtain the pseudo-annotations predictions of each worker on its unannotated instances using Eq.~\ref{annpred} 
		
		\STATE \textbf{Repeat:}
	   \STATE for each pseudo-annotation, calculate its confidence score according to Eq.~\ref{entropy}
	   \STATE for each class $c$, calculate the corresponding selection number $M_c$ according to Eq.~\ref{select_num}
		\STATE select the $M_c$ most confident pseudo-annotations within each class
		\STATE add the selected pseudo-annotations into the training data and retrain the crowdlayer model
		\STATE \textbf{Until expected performance reached}	
	\end{algorithmic}
\end{algorithm}

\noindent{\bf Confidence} Confidence is a commonly used measure in self-training, which measures how confident the prediction of the current model is for some instances. Using $\hat{p}(\overline{y}^r)$ defined in Eq.~\ref{annpred} to denote the pseudo-annotations probability of worker $r$ on some unannotated instance $x$, we propose to use entropy to measure its confidence:
\begin{equation}
	entropy(\overline{y}^r) = -\sum_{c=1}^{C}\hat{p}(\overline{y}^r) \cdot \log{\hat{p}(\overline{y}^r)} \label{entropy}
\end{equation}
The pseudo-annotations with lower entropy values are considered to be more confident and more likely to be correct. Based on traditional self-training motivation, pseudo-annotations with the least entropy values should be selected as authentic ones. However, as we discussed in the introduction, without taking the class-imbalance issue into account, the learning algorithm would be biased towards selecting majority class annotations and ignore the minority annotations. More seriously, this bias can accumulate throughout the training process, which will inevitably damage the performance. In the following, we propose our distribution aware confidence measure.

\noindent{\bf Distribution Aware Confidence} Resampling is a common strategy for addressing the class-imbalance problem. It intuitively oversamples the majority classes or undersamples the minority classes to avoid the dominant effect of majority data. In this paper, we adopt the resampling strategy within each class, i.e., the $M_c$ most confident pseudo-annotations for each class $c\in \{1,\cdots,C\}$ are selected: 
\begin{align} 
	M_c = {t_c} \cdot M, \;\;\;\;\;\;\;\;
	\sum_{c=1}^{C} t_c = 1.
	\label{select_num}
\end{align}

Here $M$ denotes the total number of selected pseudo-annotations within each iteration, which is a hyperparameter set by the users. ${t_c}$ denotes the normalized fraction coefficient of class $c$, which is inversely proportional to the number of pseudo-annotations $N'_c$ of class $c$ among all the generated pseudo-annotations:
\begin{align}\label{prop}
	{t_c} \propto  \frac{1}{N'_c}.
\end{align}

Algorithm \ref{alg:algorithm} summarizes the main steps of the Self-Crowd approach. We iteratively predict the unobserved annotations and add the most confident ones into the training data. Those pseudo-annotations with lower entropy values and rebalancing the annotation distribution are selected according to Eq.~\ref{select_num}-~\ref{prop}. Then the learning model is retrained on the combination of observed and pseudo-annotations. This process repeats until the expected performance is reached. 

\begin{figure*}[htbp]
	\centering
	\begin{minipage}[t]{\textwidth}
		\centering
		\includegraphics[width=0.75\textwidth]{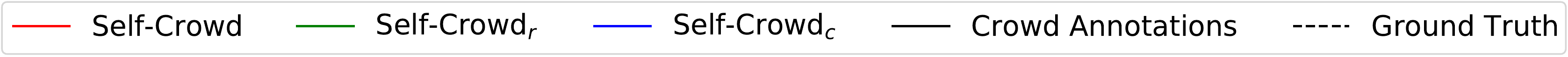}
	\end{minipage}
	\begin{minipage}[t]{\textwidth}
		\centering
		\subfigure[Test Accuracy]{
			\centering
			\includegraphics[width=0.35\textwidth]{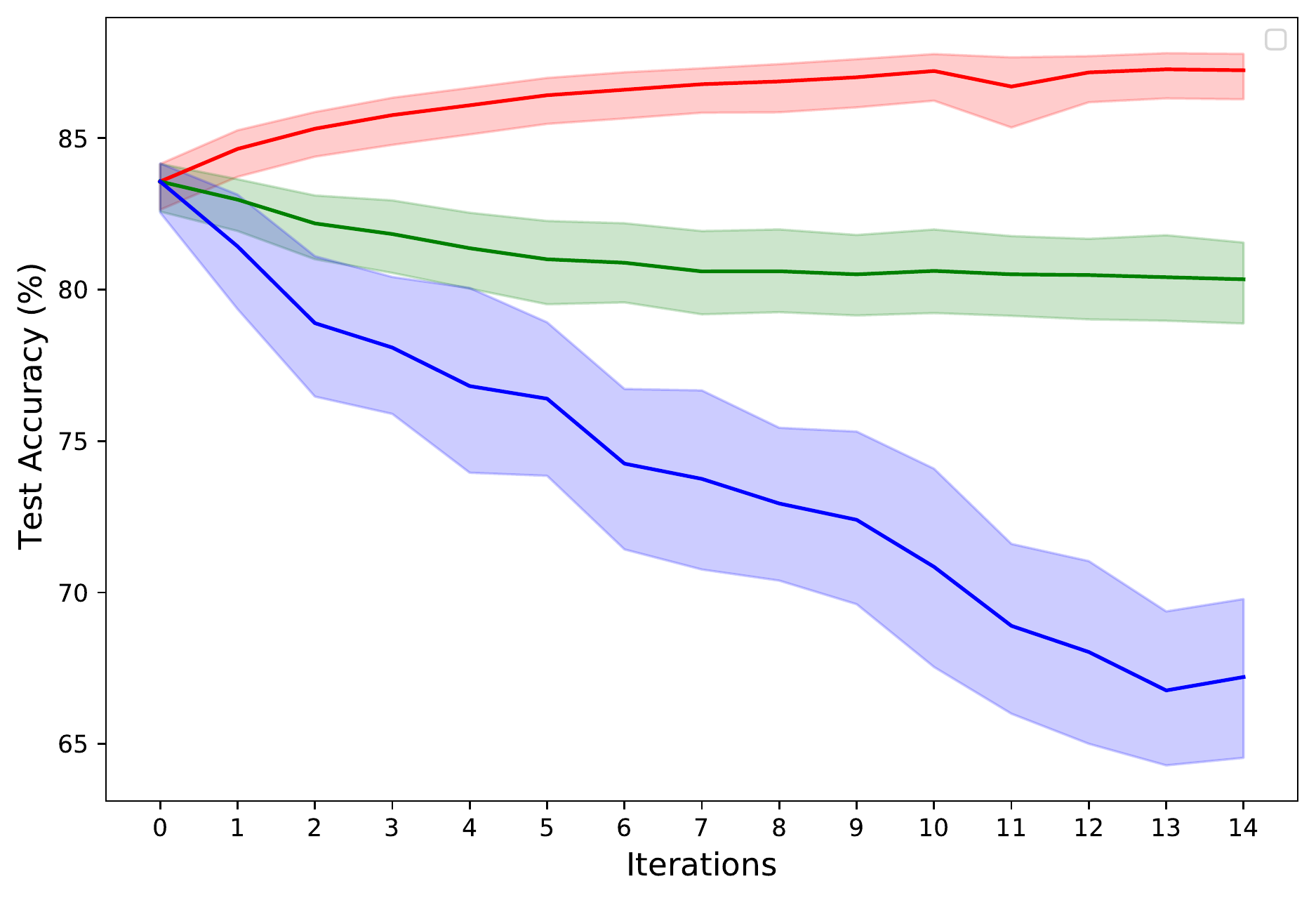}
		}
		\subfigure[$R$ of Pseudo-Annotations]{
			\centering
			\includegraphics[width=0.3\textwidth]{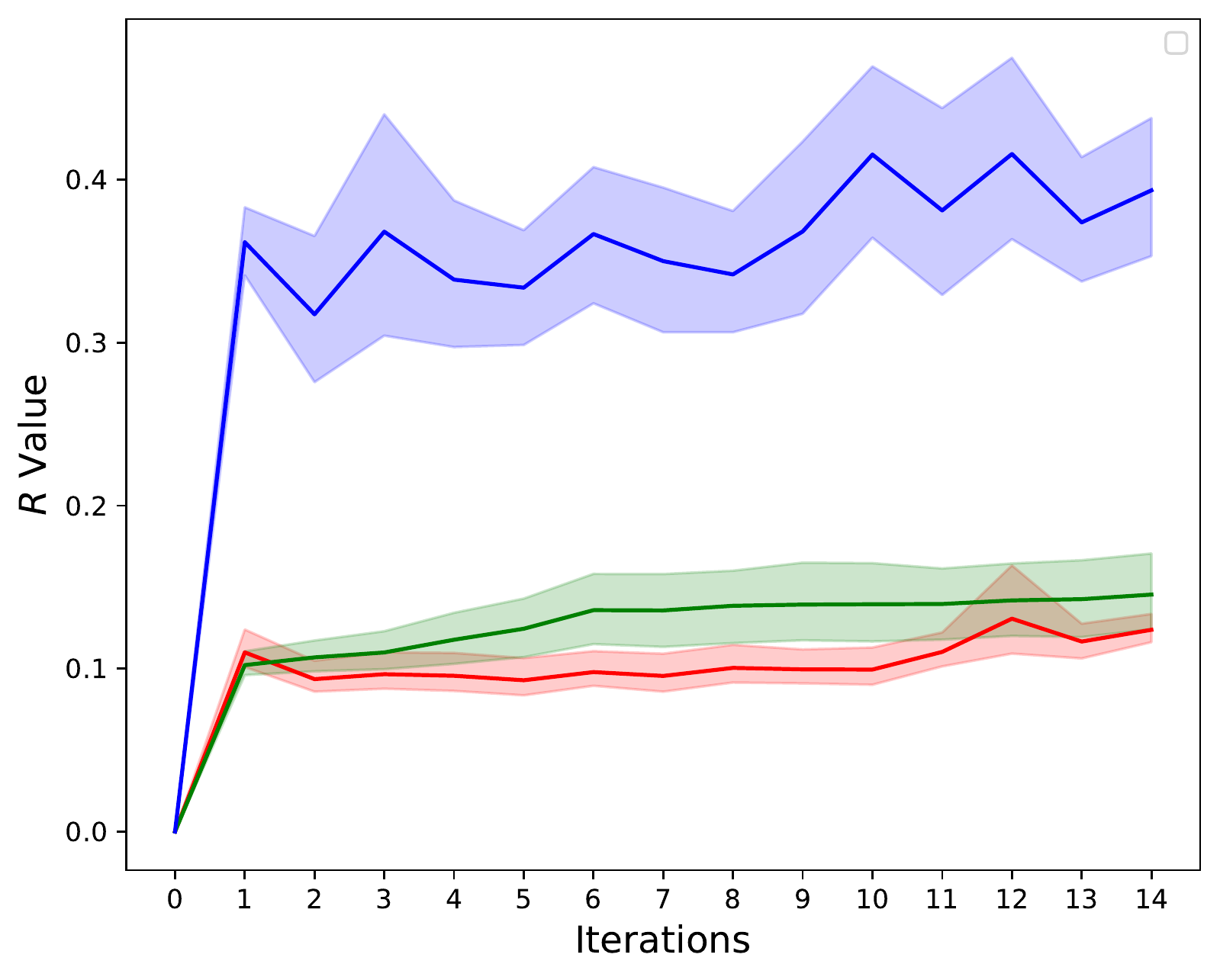}
		}
		\subfigure[$R$ of Combined Annotations]{
			\centering
			\includegraphics[width=0.3\textwidth]{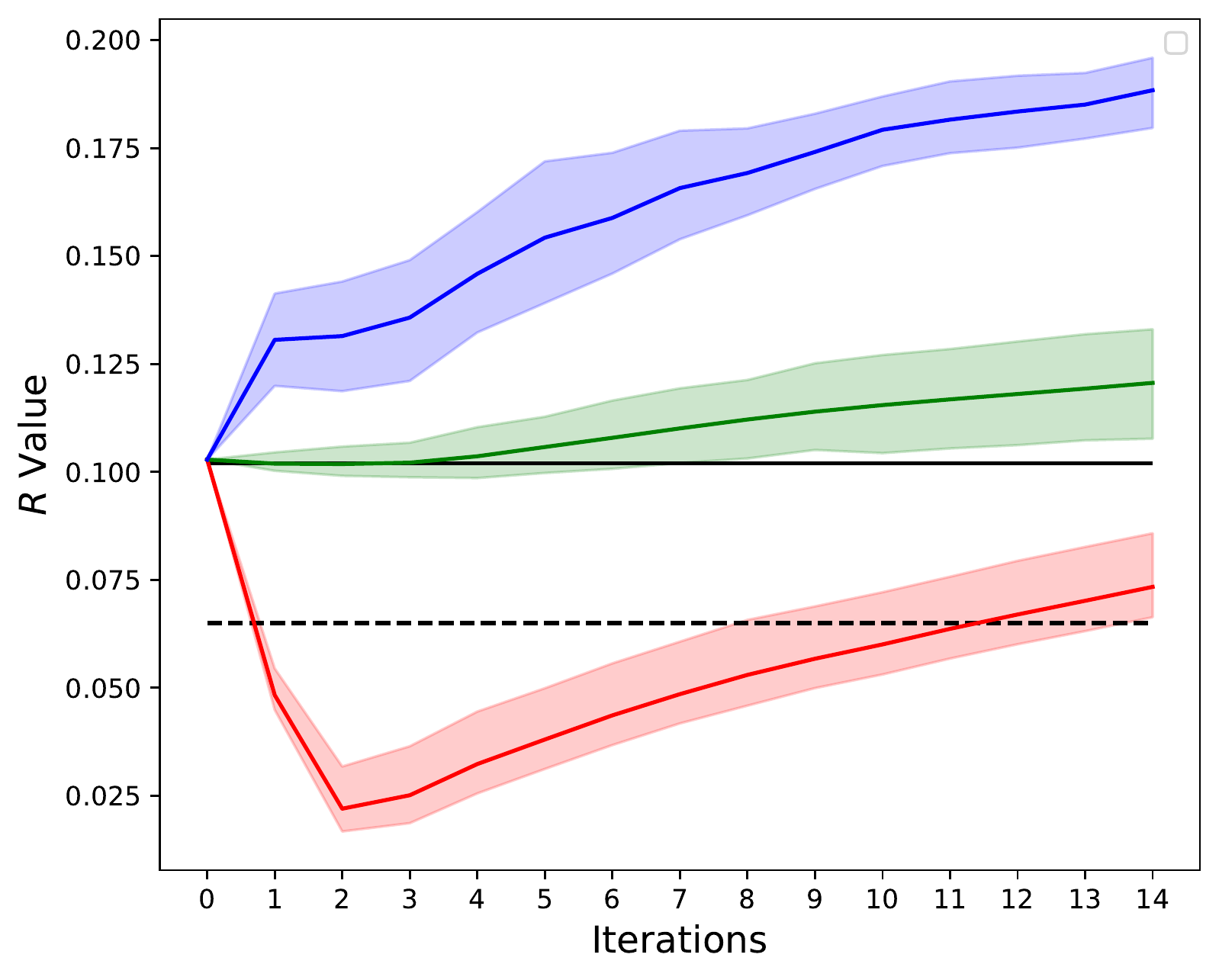}
		}
	\end{minipage}
	
	\caption{Comparison of {\it Self-Crowd}, {\it {Self-Crowd}$_{r}$} and {\it {Self-Crowd}$_{c}$} from three different perspectives.}
	\label{fig:comparsions}
\end{figure*}

\section{Experiments}
\subsection{Settings}
\noindent\textbf{Dataset} We conduct experiments on LabelMe\cite{labelme}, a real-world crowdsourcing image classification dataset. LabelMe consists of respectively $1000$ and $1688$  training and testing images concerning $8$ classes: {\it "highway", "inside city", "tall building", "street", "forest", "coast", "mountain" and "open country"}. The authors distributed the training images to $59$ crowd workers through the Amazon Mechanical Turk (AMT) platform, and got on average $2.547$ annotations for each image.
The accuracy of each worker ranges from $0$ to $100\%$, with mean accuracy and standard deviation $69.2\% \pm 18.1\%$. As shown in Figure ~\ref{fig:hist_classes}, the LabelMe dataset is imbalanced with regard to both the ground truth labels and collected crowds annotations. 

\noindent\textbf{Network and Optimization} For a fair comparison, we implement the methods following the setting in \cite{crowdlayer}. Specifically, we use the pretrained CNN layers of the VGG-16 deep neural network~\cite{vgg16} with one fully connected layer with 128 units and ReLU activations and one output layer on top. Besides, $50\%$ random dropout is exploited. The training is conducted by using Adam optimizer\cite{adam} for $25$ epochs with batch size $512$ and a learning rate of $0.001$. $L_2$ weight decay regularization is used on all layers with $\lambda =0.0005$. 

\noindent\textbf{Baselines} To assess the performance of the proposed approach, we conduct comparisons for the following implementations: 

\noindent{\it {Self-Crowd}$_{r}$}: which randomly selects the pseudo-annotations and is used for training.

\noindent{\it {Self-Crowd}$_{c}$}: which selects the most confident pseudo-annotations with the least entropy values according to Eq.~\ref{entropy} and used for training without considering class-imbalance.

\noindent{\it Self-Crowd}: which selects the pseudo-annotations taking into account class-imbalance issue according to Eq.~\ref{entropy}-~\ref{select_num}.

We examine the classification accuracy on the test images. To avoid the influence of randomness, we repeat the experiments for 20 times and report the average results.

%

\subsection{Results}

Figure \ref{fig:comparsions} (a) shows the test accuracy of compared methods on the LabelMe dataset as the self-training process iterates. Here results for $14$ iterations are recorded, and in each iteration $10,000$ pseudo-annotations are selected without replacement. As we can see, the test accuracy of {\it {Self-Crowd}$_{c}$} and {\it {Self-Crowd}$_{r}$} decrease rapidly as the self-training proceeds. In contrast, {\it Self-Crowd} stably improves.

To examine what happened during the learning procedure, we define one class-imbalance ratio $R$ as following: 
\begin{equation}
	R = \frac{N_{max} - N_{min}}{N_{anno}} \label{Imbalance ratio}
\end{equation}
Here $N_{min}$, $N_{max}$ respectively denote the number of generated annotations for the most frequent class and the least frequent class, $N_{anno}$ denotes the total number of generated annotations over all classes. It can be seen that $R$ ranges in $[0,1]$, with smaller value meaning more balanced annotations.

We record the variation of $R$ for the pseudo-annotations selected by the three methods during self-training in Figure~\ref{fig:comparsions} (b). It can be seen that the $R$ value of {\it {Self-Crowd}$_{c}$} increases rapidly, indicating that the confidence based measure mostly selects the majority class pseudo-annotations, leading to severely imbalanced annotation distribution, which in turn badly hurt the learning performance as shown in Figure~\ref{fig:comparsions} (a). The random selection strategy is much better than confidence based measure but still biased by the imbalance issue. The proposed distribution aware strategy is more robust and achieves improved performance.

Combing the original observed annotations and the selected pseudo-annotations, Figure~\ref{fig:comparsions} (c) shows the $R$ value variation on the combined annotations. The solid and dashed black line respectively represents the $R$ value of original observed annotations and the ground truth labels. It can be seen that our proposed method greatly alleviates the class-imbalance issue during learning whereas the random and confidence based selection measure always leads to more imbalanced annotations. This explains the performance decline of {\it {Self-Crowd}$_{c}$} and {\it {Self-Crowd}$_{r}$}.

\subsection{Various Sparsity Level Study}

\begin{figure}[H]
	\centering
	\includegraphics[width=0.4\textwidth]{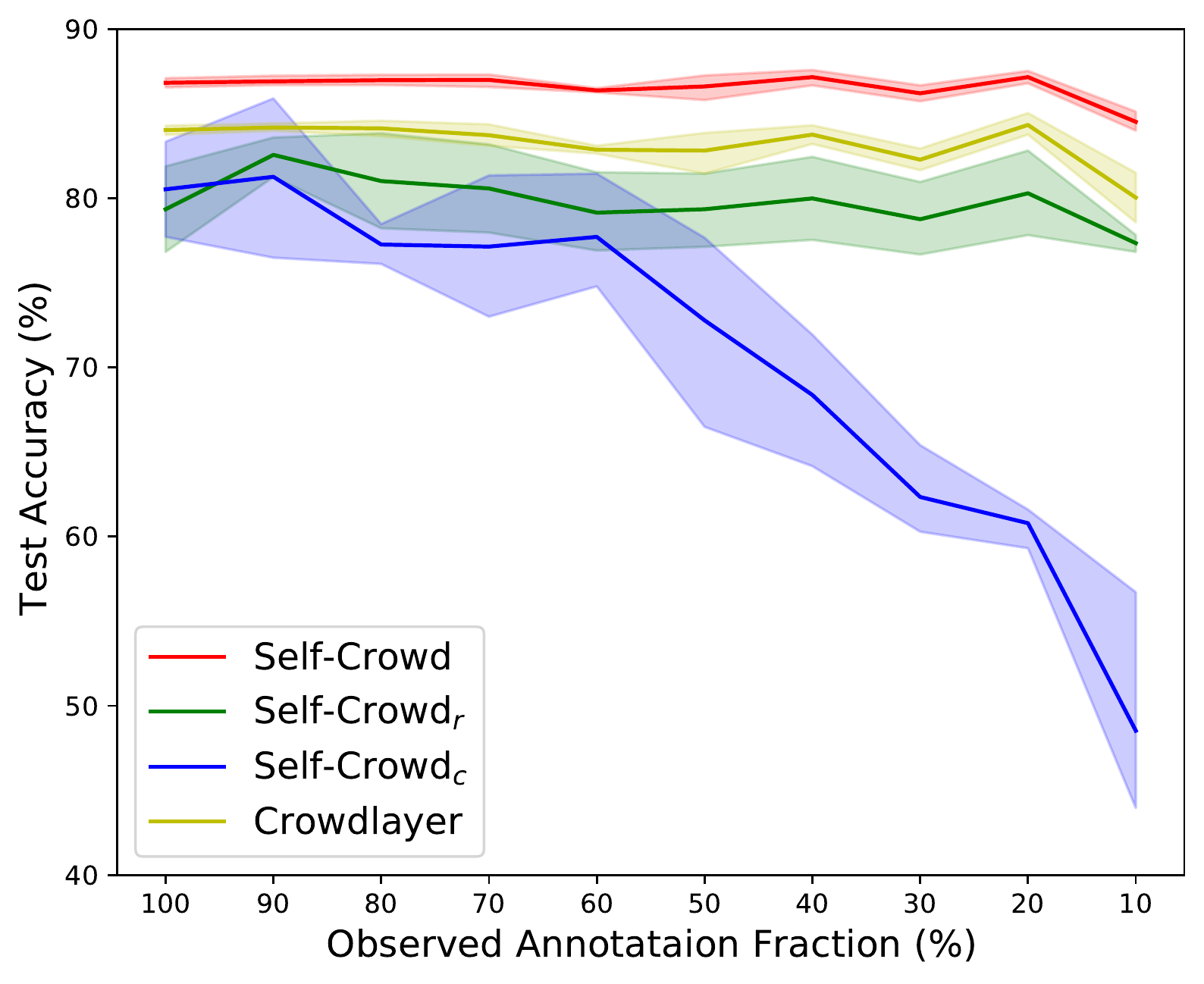}
	\caption{Test accuracy with different sparsity levels.}
	\label{fig:sparsity}
\end{figure}

To examine the effectiveness of our approach with different sparsity levels, we remove fractions of the original observed annotations for LabelMe and conduct experiment. Specifically, we remove $p$ fractions of the observed annotations with $p$ ranges from $0\%$ to $90\%$ in a uniformly random manner. To alleviate the effect of randomness, we repeat each experiments for 5 times and report the average results. For the self-training process, 5 iterations are conducted with each $10,000$ pseudo-annotations selected within each iteration. Figure ~\ref{fig:sparsity} shows the results. 

The yellow line represents the test accuracy when only the observed annotations are used for training without self-training, i.e., $t=0$. It can be seen our approach always achieves the best and most stable performance. However, the confidence based approach {\it Self-Crowd$_{c}$} decreases rapidly as the observation annotations decrease, and {\it Self-Crowd$_{r}$} performs stably but worse than the crowdlayer baseline.

\section{Conclusion}

In this paper, we propose a self-training based method Self-Crowd to deal with the sparsity and class-imbalance issue in crowdsourcing learning. To combat the selection bias towards majority class annotations, we propose a distribution aware confidence measure to select the most confident pseudo-annotations and rebalance the annotation distribution. Experiments on a real-world crowdsourcing dataset show the effectiveness of our approach. As a primary attempt to sparse and imbalance crowdsourcing learning, the proposed method can be extended by combining with sophisticated deep crowdsourcing learning models and selection measures.

%

\appendix

\bibliographystyle{named}
\bibliography{Self-Crowd}

\end{document}